\pdfoutput=1
\documentclass[Alon2,ChapterTOCs,singlecolor,11pt]{Alon}
\usepackage{fixltx2e,fix-cm}
\usepackage[english]{babel}
\usepackage{epigrafica}
\usepackage[LGR,OT1]{fontenc}
\usepackage{mathptmx,txfonts}
\usepackage{lmodern}
\usepackage{amssymb}
\usepackage{txfonts} % or any other font package
\usepackage{graphicx}
\usepackage{subfigure}
\usepackage{makeidx}
\usepackage{multicol}
\usepackage{booktabs}
\usepackage{multirow}
\usepackage{enumitem}
\usepackage{comment}

\usepackage{cite}

\usepackage[table,xcdraw]{xcolor}

\usepackage{textcomp}
\usepackage{comment}
\usepackage{enumerate}
\usepackage{caption}

\usepackage{rotating}

\usepackage{changepage}% for text width to include margin

\makeindex

\usepackage{version}

\usepackage{lipsum} % To generate some text for the article via \lipsum

\usepackage{tikz,xcolor}
\usepackage{url} 

\definecolor{lime}{HTML}{A6CE39}
\DeclareRobustCommand{\orcidicon}{%
	\begin{tikzpicture}
	\draw[lime, fill=lime] (0,0) 
	circle [radius=0.16] 
	node[white] {{\fontfamily{qag}\selectfont \tiny ID}};
	\draw[white, fill=white] (-0.0625,0.095) 
	circle [radius=0.007];
	\end{tikzpicture}
	\hspace{-2mm}
}

\foreach \x in {A, ..., Z}{%
	\expandafter\xdef\csname orcid\x\endcsname{\noexpand\href{https://orcid.org/\csname orcidauthor\x\endcsname}{\noexpand\orcidicon}}
}

\usepackage{epstopdf}% To incorporate .eps illustrations using PDFLaTeX, etc.
\usepackage[caption=false]{subfig}% Support for small, `sub' figures and tables
\renewcommand\bibfont{\fontsize{10}{12}\selectfont}% To set the list of references in 10 point font using natbib.sty

\title{Taylor and Francis Book Chapter}
\begin{document}

% \frontmatter

\begingroup
\pagestyle{empty}
\cleardoublepage
\endgroup

\maketitle %Computer Vision and Image Analysis for Industry 4.0

%%%Placeholder for front matter

%\halftitle

%\booktitle Computer Vision and Image Analysis for Industry 4.0

%\locpage

\include{frontmatter/dedication}
%\cleardoublepage
\setcounter{page}{1} %previous pages reserved for frontmatter to be added later
\tableofcontents
%\listoffigures
%\listoftables
%\include{frontmatter/foreword}
%\include{frontmatter/preface}
%\include{frontmatter/contributor}
%\include{frontmatter/authorbio}

% \mainmatter

% Clear Blank Page

\let\cleardoublepage\clearpage

%\part{This is a Part}

% new add
\newcommand*\rot{\rotatebox{90}}
\newcommand*\OK{\ding{51}}

\chapterauthor{M. M. Akash*}{visie.tech, Dhaka, Bangladesh\\
Email: akash18rayhan@gmail.com, *Corresponding Author}
\chapterauthor{Rahul Deb Mohalder}{Khulna University, Khulna, Bangladesh\\
Email: rahul@ku.ac.bd}
\chapterauthor{Md. Al Mamun Khan}{visie.tech, Dhaka, Bangladesh\\
Email: almamunkhan04@gmail.com}
\chapterauthor{Laboni Paul}{Khulna University, Khulna, Bangladesh\\
Email: laboni1124@cseku.ac.bd}
\chapterauthor{Ferdous Bin Ali}{visie.tech, Dhaka, Bangladesh\\
Email: hridoyferdous@yahoo.com}

\chapter{Yoga Pose Classification Using Transfer Learning}

\chaptermark{Yoga Pose Classification Using Transfer Learning}

\chapterinitial {Y}oga has recently become an essential aspect of human existence for maintaining a healthy body and mind. People find it tough to devote time to the gym for workouts as their lives get more hectic and they work from home. This kind of human pose estimation is one of the notable problems as it has to deal with locating body key points or joints. Yoga-82, a benchmark dataset for large-scale yoga pose recognition with 82 classes, has challenging positions that could make precise annotations impossible. We have used VGG-16, ResNet-50, ResNet-101, and DenseNet-121 and finetuned them in different ways to get better results. We also used Neural Architecture Search to add more layers on top of this pre-trained architecture. The experimental result shows the best performance of DenseNet-121 having the top-1 accuracy of 85\% and top-5 accuracy of 96\% outperforming the current state-of-the-art result.

\section{Introduction}
Human Activity Recognition (HAR) is a field in computer vision that involves identifying and classifying human activities from sensor data or video streams. This can be done using various techniques, such as machine learning algorithms, pattern recognition, and computer vision.

Yoga poses are a specific type of human activity that can be recognized using HAR techniques. By analyzing sensor data or video streams of a person performing yoga, a HAR system can identify and classify the specific yoga poses being performed. This can be useful for a variety of applications, such as yoga instruction, and injury prevention. HAR system could be used to track a person’s progress as they perform a series of yoga poses, providing feedback and guidance on proper form and technique. It could also be used to identify any incorrect or potentially harmful movements, helping to prevent injuries. 

Machine learning is being used to help people exercise more effectively through the use of virtual personal trainers. These are systems that use machine learning algorithms to provide personalized exercise routines and coaching based on the user's goals, fitness level, and other factors. By analyzing data such as heart rate, body composition, and exercise performance, these systems can adapt and modify the user's routine in real-time to optimize their results. Wearable technology such as fitness trackers and smartwatches are increasingly using machine learning to improve their accuracy and effectiveness. For example, some fitness trackers use machine learning algorithms to analyze data such as steps taken, calories burned, and heart rate variability to provide more accurate and personalized feedback to the user.

During the COVID-19 pandemic, Yoga is a popular practice that has been shown to reduce stress, anxiety, and depression. However, to get the maximum benefits of yoga, it is crucial to perform the poses correctly, which requires access to a yoga trainer. Unfortunately, not everyone has access to a trainer. For these concerned people, a software application can play the role of an instructor. Traditional pose estimation methods for recognizing and correcting poses are at a disadvantage here \cite{1_verma2020yoga} because of the diversity of poses on human bodies. Therefore, this paper proposes a novel solution, a classification system that captures the similarities and differences between poses. Instead of producing body key points for human subjects, this method recognizes the overall pose and classifies it. The proposed approach can improve access to yoga for individuals who do not have access to a trainer and enhance the benefits of yoga for all practitioners.

\subsection{Our Contribution:}
The key contributions of this research are:

\begin{itemize}
  \item 	We have experimented with different types of image preprocessing techniques. 
  \item 	We have used transfer learning in different stages like using fully imagenet weight, finetuning different layers of the model, and finetuning the whole model.  
 \item We have defined the network using Random Search for architecture.

\end{itemize}

The paper is organized as follows, the very next section (section \ref{sec_literature_survey}) contains the previous work description. After that, we introduced our methodology in section \ref{sec_methodology} followed by result analysis in section \ref{sec_result_analysis} and the conclusion and future work in section \ref{sec_conclusion_and_future_work}.

\section{Related Work}
\label{sec_literature_survey}
Chen\cite {2_chen2020monocular} conducted various deep learning (DL) based techniques to deduce the human pose estimation from images. The usage of Deep learning for human pose estimation was established when AlexNet used CNN  and achieved state-of-the-art results. The researchers pointed out that there are now a variety of pose estimate techniques, including pixel-level analysis, body joint point mapping, body-based and model-based human body estimation, and heat-map mapping.

Faisal \cite{3_faisal2019monitoring} continued on Chen's body-joint estimation research \cite{2_chen2020monocular} and mentioned the current state-of-the-art, where they repeated that the body-joint points are identified by employing different gyroscope and joint angle methods for determining joint point angle, as well as multiple sensor fusion.

The authors of \cite{4_boualia2019pose} repeated Faisal and Chen's findings from \cite{2_chen2020monocular}, \cite{3_faisal2019monitoring} by estimating human activity based on poses. It should be noted that prominent posture estimation techniques include template-based, generative model-based, and discriminative model-based techniques. These techniques were later verified in research on human pose estimation published in \cite{5_chowdhury2020hactnet}. Yoga has been more well-liked among people because of its believed benefits of good health and fitness. Nagalakshmi examined the effects of yoga and discovered that remote yoga activities have become significantly more popular since the COVID outbreak, in addition to alleviating musculoskeletal pain and also inspiring a healthy lifestyle \cite{6_nagalakshmi2021analysis}.

For this goal, Agarwal \cite{7_agrawal2020implementation} experimented with various Machine Learning (ML) methods. They created a dataset of 5,500 photos for ten distinct yoga postures and then they used the open pose estimation algorithm to create the skeleton images. On their dataset, they acknowledged that random forest classifiers performed better. Liaqat et al. \cite{8_9343347} created a hybrid posture recognition approach by combining deep neural networks with conventional machine learning techniques. The weight learned by the deep learning model and the prediction from the conventional model are combined to produce the final class prediction. In a different paper \cite{9_kumar2020yoga}, the authors first extracted key points using OpenPose, then classified the data using a hybrid CNN-LSTM layer. The researchers on this team worked for sitting pose estimation \cite{10_kulikajevas2021detection} and concluded a Deep Recurrent Hierarchical Network based on MobileNetV2 and gained 91.47\% accuracy. Similarly, the use of SVM and boosting for stance detection is addressed in \cite{11_panigrahy2021detection}, and Nagalakshmi's work \cite{12_nagalakshmi2021classification} specifically focused on yoga.

 Keeping pace with deep neural networks pose estimation models are getting more effective day by day. The mostly accurate keypoints and skeleton annotations are the strength of state-of-the-art pose estimation models. Pose estimation models are widely used in human surveillance \cite{13_chen2018shpd}, and human activity recognition \cite{14_holte2012human} like computer vision problems. So far we are working with yoga poses as a result we can say pose estimation models can make way for this problem. However, the authors \cite{1_verma2020yoga} opposed the idea of using pose estimation models for the yoga-82 dataset. They showcased some yoga poses that seem too complex from a single point of view because the human body can handle heavy diversity. With the change in image resolutions, it gets worse. Considering these factors they stated that pose models extracted keypoints and skeletons tend to have false and wrong annotations. As a result, they proposed the idea of hierarchical pose classification. They proposed hierarchical labeling where categories are created by the variations in body postures like standing, sitting, etc. And then subcategories are created from those categories.

\section{Methodology}
\label{sec_methodology}

\begin{figure}
	\centering
	\includegraphics[ width=0.95\textwidth]{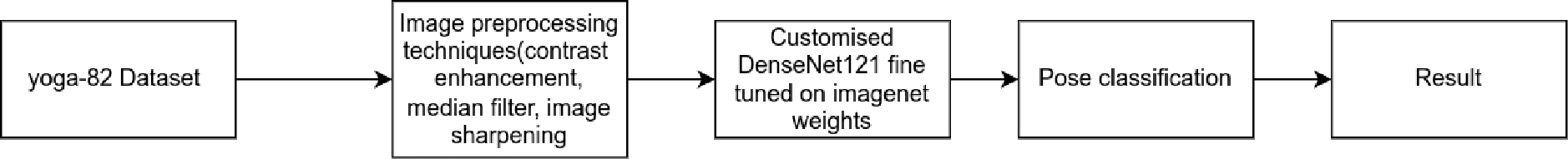}
	\caption{Our Proposed Workflow Diagram.}
	\label{fig:workflow}
\end{figure}

The very first step is to acquire the hierarchical poses (Figure \ref{fig:workflow}) of the yoga-82 dataset. Before feeding the images into the model we applied some preprocessing techniques to enhance image quality such as contrast enhancement, median filter, and image sharpening. The next involves a deep learning model architecture as a backbone here we have selected DenseNet-121 and added some dense blocks on top of that architecture at the very end adding the classification layer with 82 nodes. By using pre-trained weights on the ImageNet dataset we have fine-tuned our model on the specific yoga-82 dataset. While training the model we applied data augmentation techniques like rotation, zooming, and shearing which helped the model generate more generalized predictions.

\subsection{Dataset}
\begin{figure}
	\centering
	\includegraphics[ width=1.0\textwidth, height=1.2\textwidth]{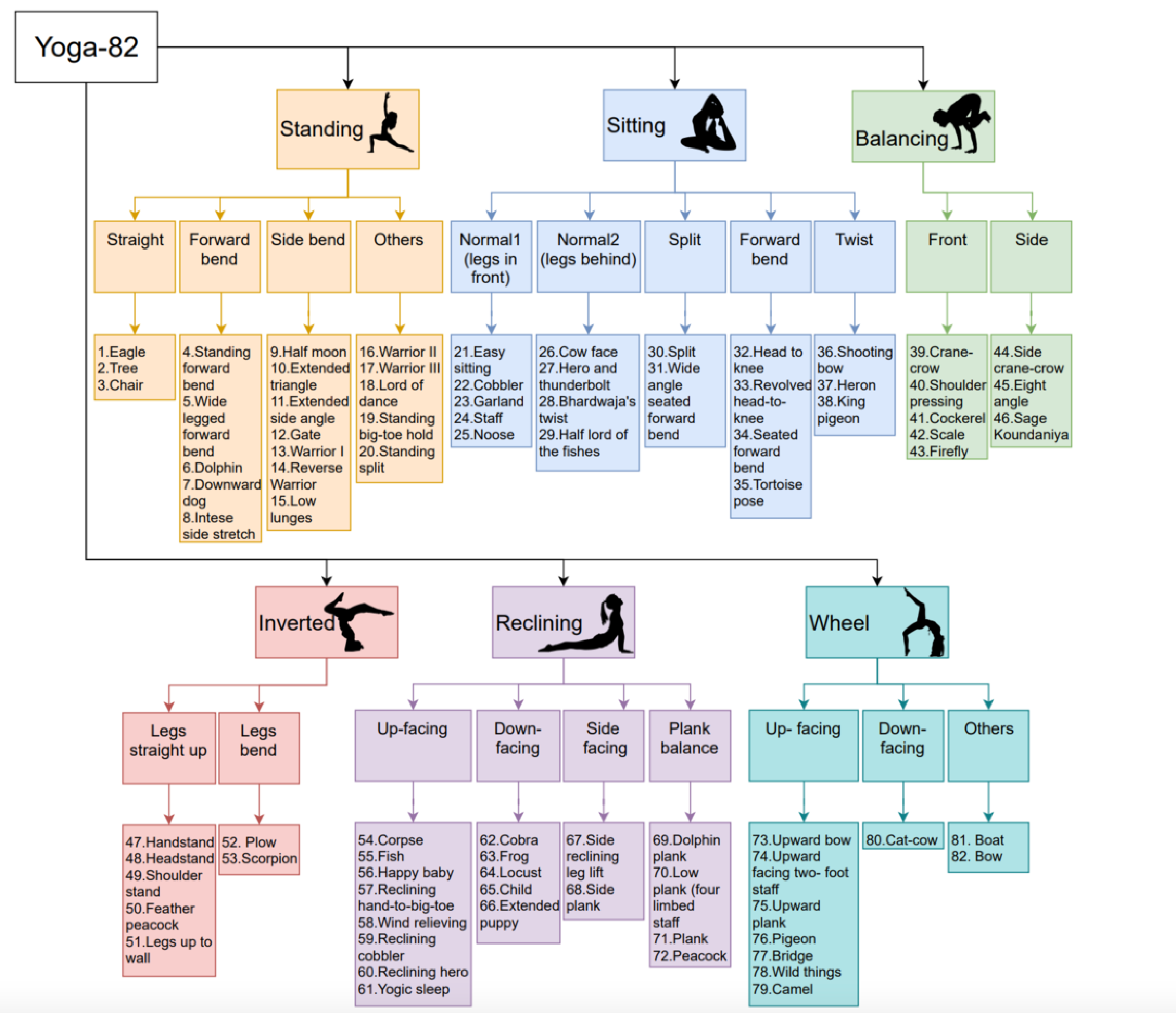}
	\caption{Yoga-82 Dataset Description \cite{1_verma2020yoga}.}
	\label{fig:dataset}
\end{figure}

Yoga-82 \cite{1_verma2020yoga} contains 21,009 train images and test images 7,469 images with multilevel class hierarchy depending on the variations of pose. It has 5 classes, standing, sitting, balancing, inverted, reclining, and wheel as a base class. They are divided into 20 subclasses. In the third level, they are divided into 82 classes. We have worked with the third level of annotation. In Figure \ref{fig:dataset} we showed the Yoga-82 dataset description view.

\subsection{Data Processing}
We followed these data processing techniques to process our data.

\subsubsection{Image Enhancement}
From the PIL ImageEnhance class an object Contrast is instantiated to apply contrast on images. The concept of image contrast is different from brightness. Brightness makes the whole image brighter or darker. On the other hand, the contrast makes the light areas lighter and the dark areas darker. The lighter or darker amount depends on the factor. 

\begin{figure}
	\centering
	\includegraphics[ width=0.95\textwidth, height=0.25\textwidth]{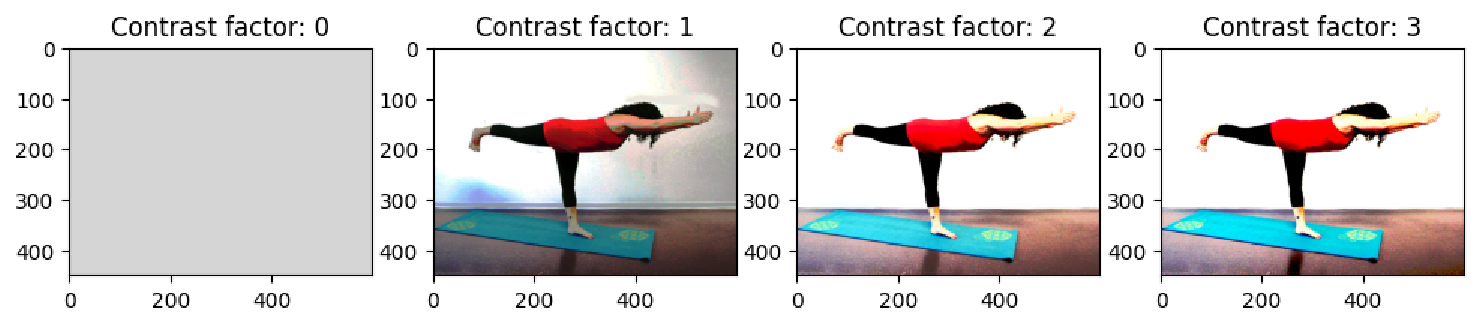}
	\caption{Effect of the contrast factor.}
	\label{fig:ie}
\end{figure}

Increasing contrast makes different parts of the body more visible. When contrast is increased in an image certain areas like body joints highlight some patterns, and muscle appears more defined. Eventually, it makes it easier for the model to extract better features (Figure \ref{fig:ie}).

\subsubsection{Median Filter}
\begin{figure}
	\centering
	\includegraphics[ width=0.95\textwidth, height=0.25\textwidth]{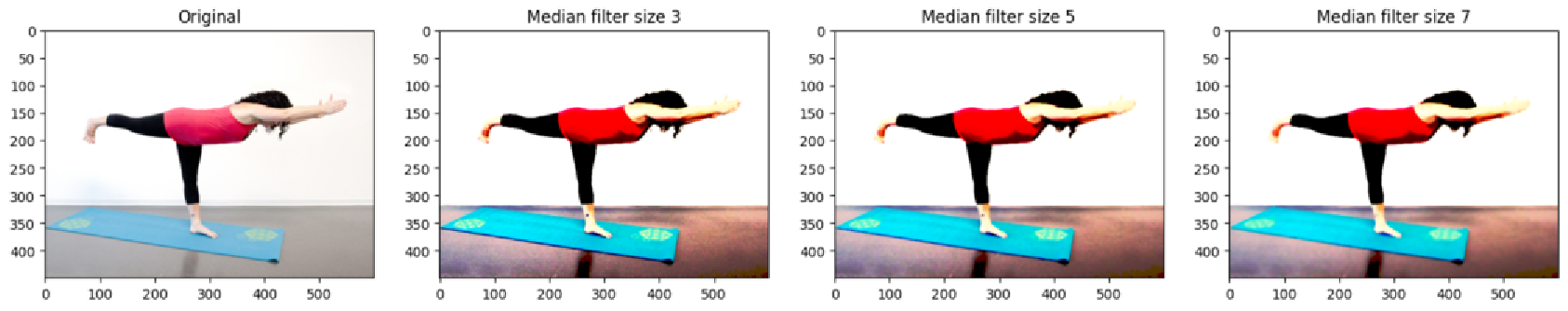}
	\caption{Effect of the median filter.}
	\label{fig:mf}
\end{figure}

When we increase the contrast in an image, we are increasing the difference between the light and dark areas of the image. This also leads to an increase in noise. To remove the noise on images we used a median filter (Figure \ref{fig:mf}). The median filter calculates the median of the pixel intensities surrounding the center pixel in a $n \times n$ kernel. The kernel we chose is $3 \times 3$.

\subsubsection{Image Sharpening}
\begin{figure}
	\centering
	\includegraphics[ width=1.0\textwidth, height=0.25\textwidth]{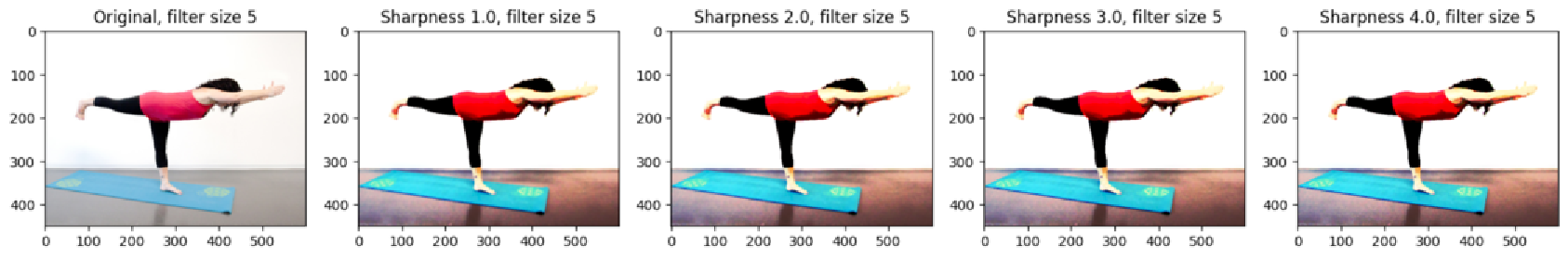}
	\caption{Effect of the image sharpening.}
	\label{fig:if}
\end{figure}

The median filter also results in a slightly blurred image and some loss of detail and edge sharpness. Applying a sharpening filter to an image increases the contrast between adjacent pixels and enhances the edges in the image, which helps to reduce the appearance of blurriness in the image (Figure \ref{fig:if}).

\subsection{Transfer Learning Models}
\begin{figure}
	\centering
	\includegraphics[ width=0.85\textwidth, height=0.5\textwidth]{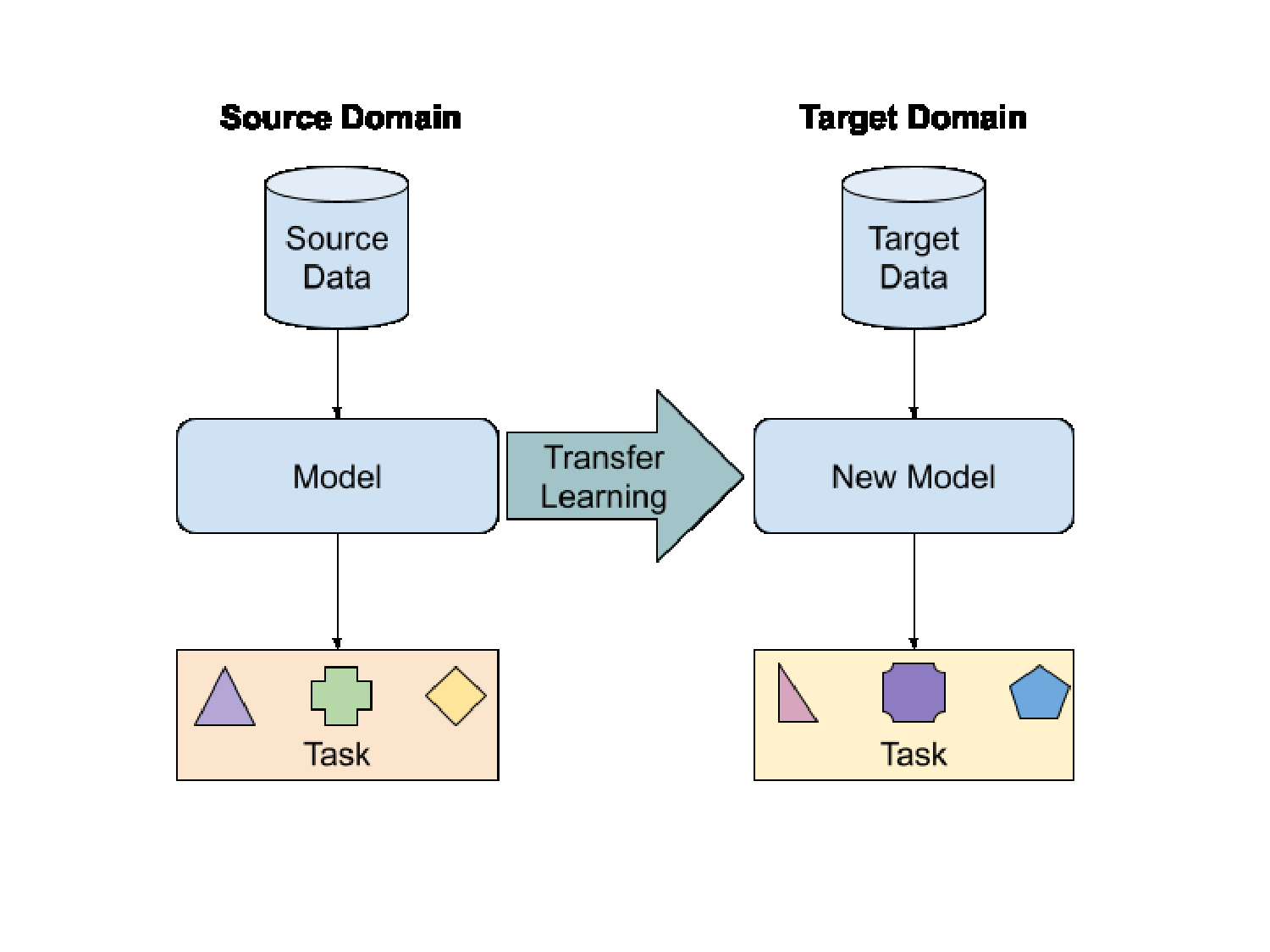}
	\caption{Transfer Learning Working Process.}
	\label{fig:tf_work}
\end{figure}

Transfer learning is a machine learning technique that has been successfully applied to many classification tasks. It involves leveraging a pre-trained model to improve the accuracy and efficiency of a new classification task (Figure \ref{fig:tf_work}).

Consider X be the set of input data, y be the set of the output labels, and f be a function that represents the pre-trained model. So f is a function that maps input X to output Y, i.e., Y = f(X). The model f has learned some underlying patterns and features that can be reused for the new classification task.

Now, let X' be the input data for the new classification task, and y' be the output labels. The goal is to train a new model g that can accurately classify the input data X' into the output labels y'. The new model g can be represented as a function that maps input X' to output Y', i.e., Y' = g(X').

To perform transfer learning, the weights of the pre-trained model f are used to initialize the weights of the new model g. This reduces the training time and enables the model to learn the specific features relevant to the new classification task. The fine-tuning process involves updating the weights of the last few layers of the pre-trained model f while keeping the earlier layers frozen. This enables the new model g to learn the specific features relevant to the new classification task while retaining the general features learned by the pre-trained model f.

Mathematically, the fine-tuning process involves minimizing the loss function between the predicted output of g and the actual output of the new classification task dataset. The loss function can be represented as follows:\\

%L(Y', Y) = - ∑ y' log(g(X')) + (1 - y') log(1 - g(X'))
%L(Y', Y) = - \sum \left[ y' \log(g(X')) + (1 - y') \log(1 - g(X')) \right]

$L(Y', Y) = - \sum \left[ y' \log(g(X')) + (1 - y') \log(1 - g(X')) \right]$ \\

Where L is the loss function, y' is the predicted output of g, and y is the actual output of the new classification task dataset. The goal is to minimize the value of L to improve the accuracy of the model.

\subsection{VGG-16}
The Visual Geometry Group at the University of Oxford created VGG-16, a convolutional neural network architecture. It is a deep neural network that in 2014 attained world-leading performance on the ImageNet dataset. The VGG-16 architecture consists of convolutional layers followed by fully connected layers.

\begin{figure}
	\centering
	\includegraphics[ width=1.0\textwidth, height=0.75\textwidth]{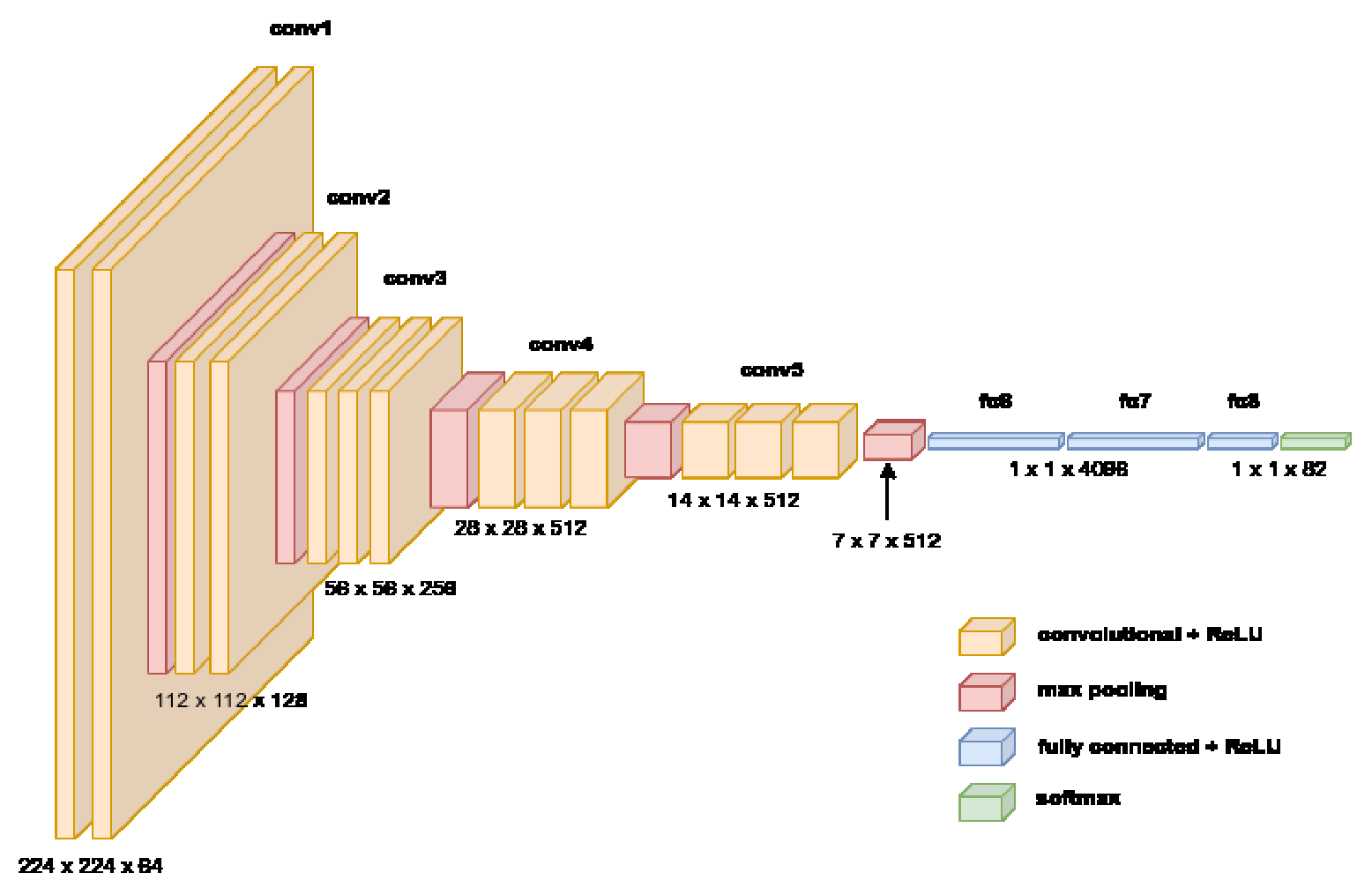}
	\caption{Modified VGG-16.}
	\label{fig:vgg16}
\end{figure}

The simplicity and clarity of the VGG-16 architecture's design are its assets. This makes it a popular choice among newcomers to deep learning, including researchers and practitioners. In addition, the VGG-16 (Figure \ref{fig:vgg16}) architecture has been utilized extensively as a pre-trained model for transfer learning, where the weights of the pre-trained network are fine-tuned for a specific task.

Overall, VGG-16 is a potent convolutional neural network architecture that has contributed significantly to the field of computer vision. Its success has prompted additional research into deep learning and created new opportunities for image classification, object recognition, and other computer vision tasks.

\subsection{ResNet-50}
In 2015, ResNet-50 joined the rest of the ResNet family of neural networks. To fix the problem of disappearing gradients, the Microsoft Research team created a convolutional neural network. The vanishing gradient problem was addressed by the use of skip connections.

\begin{figure}
	\centering
	\includegraphics[ width=1.0\textwidth, height=0.70\textwidth]{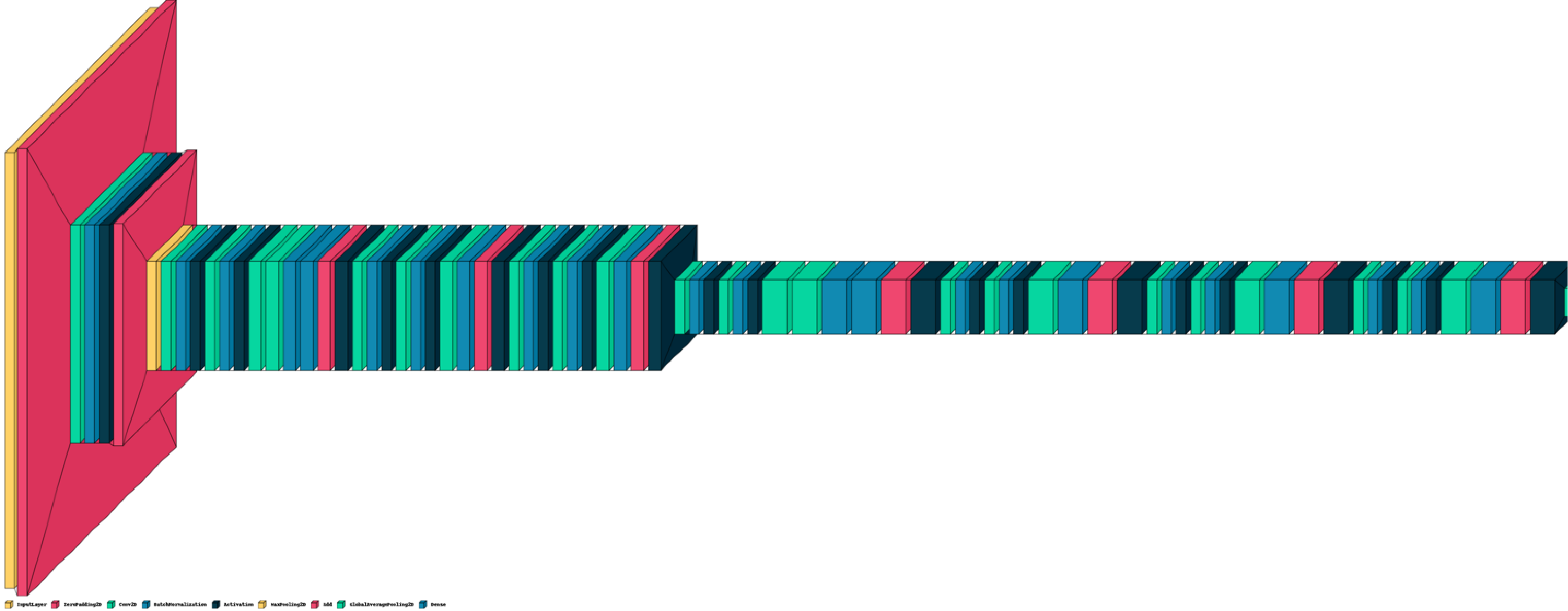}
	\caption{Modified ResNet-50.}
	\label{fig:resnet50}
\end{figure}

ResNet-50 consists of 50 layers, some of which are convolutional while others are batch normalization and others are completely linked. The first few layers are responsible for extracting low-level features from the input picture, while the latter layers are in charge of extracting and classifying higher-level characteristics (Figure \ref{fig:resnet50}). ResNet-50's capacity to accurately train incredibly deep neural networks is one of its greatest capabilities. Image categorization, object recognition, and semantic segmentation are just a few of the many computer vision tasks that have benefited from their use. 

\begin{figure}
	\centering
	\includegraphics[ width=1.0\textwidth, height=0.7\textwidth]{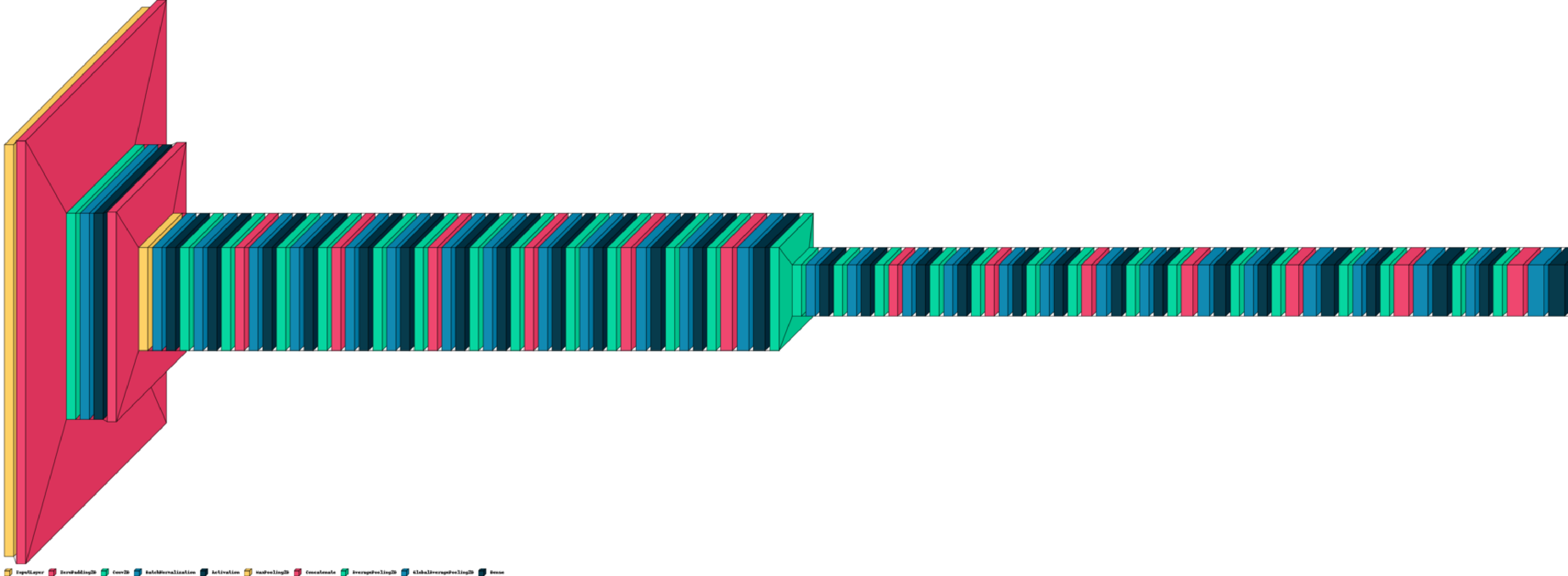}
	\caption{Modified DenseNet-121.}
	\label{fig:desnet121}
\end{figure}

\subsection{DenseNet-121}
DenseNet-121 is a convolutional neural network architecture developed by University of Toronto researchers in 2017. It is a member of the DenseNet network family, which is notable for its thick connection pattern between layers.

DenseNet-121 consists of 121 layers, which include convolutional layers, batch normalization layers, and pooling layers. The architecture is intended to capture both low-level and high-level information in the input picture, and its dense connection network aids in the prevention of overfitting and the improvement of training efficiency (Figure \ref{fig:desnet121}).
 
DenseNet-121 has shown cutting-edge performance on a variety of computer vision applications, including image classification, object identification, and semantic segmentation. It has also been demonstrated to generalize effectively to new tasks when used as a starting point for transfer learning on other datasets.

\section{Result Analysis and Discussions}
\label{sec_result_analysis}
The VGG16 network from Keras and its weights after they are pre-trained on the ImageNet dataset. Also, we removed the classifier part of this network. On ConvNet the early layers contain very generic feature maps that will be useful for our dataset even if it is very different from ImageNet, so we started with fine-tuning only the last five layers. As a result, we achieved better accuracy for VGG16 architecture. However, this idea did not perform well when we were working with ResNet-50 architecture. There are 5 stages in the model. Each stage consists of multiple convolution layers and batch normalization. In this architecture, we tried to freeze all the stages except for the last stage layers. But the result turned out to be bad compared to fine-tuning all the layers. 

It is possible that the skip connections in ResNet-50 architecture could have played a role in this difference. Skip connections in ResNet-50 allow the gradients to flow directly to the earlier layers, which can help the earlier layers learn some feature maps related to our target dataset. It also means that each layer's output may depend on the earlier layers' output. Therefore, when we fine-tune only the last stage layers, we may not be capturing the full complexity of the images.  

\begin{figure}
	\centering
	\includegraphics[ width=0.6\textwidth, height=0.45\textwidth]{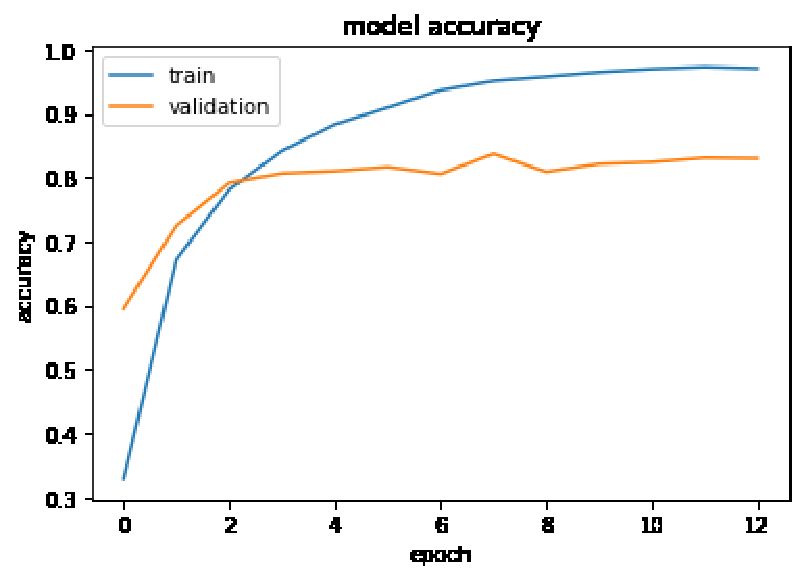}
	\caption{Classification accuracy for our modified DensNet-121.}
	\label{fig:accuracy}
\end{figure}

\begin{figure}
	\centering
	\includegraphics[ width=0.6\textwidth, height=0.45\textwidth]{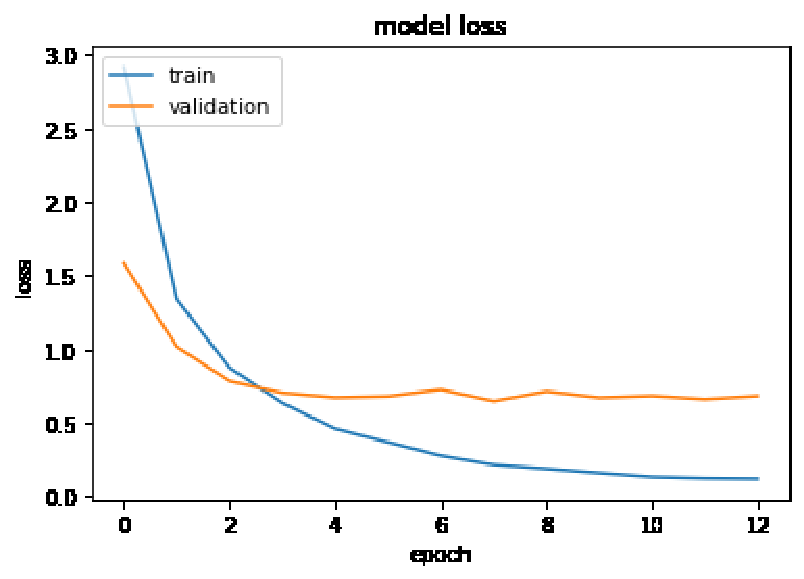}
	\caption{Classification loss for our modified DensNet-121.}
	\label{fig:loss}
\end{figure}

But we got our best performance in DenseNet-121 (Figure \ref{fig:accuracy} and Figure \ref{fig:loss}). The result in the table shows all our experiment outcomes. We have outperformed the current SOTA result. We have used four metrics to assess our classification models: accuracy, f1-score, precision, and recall. In Table \ref{tab:comparison_exp}, we have compared the results among our models. In Table \ref{tab:comparison_methos}, we have compared the best performance against the SOTA results.

Python tool packages like Keras, numpy, sci-kit learn, pandas, and Matplotlib were used to set up the project. NVIDIA 3060 GPU was used to run and train the project. The models were trained for 500 epochs with 256-person batches and a 0.01-percent learning rate with exponential decay. There was the use of Adam's planner. Hyperparameter search was done with Kears Autotuner, which had a return method that kept track of validation loss and accuracy. Models were stopped early by using early stopping with a 15-epoch tolerance number.

% \begin{figure}[ht!]
%      \begin{center}
% %
%         \subfigure[]%Caption of First Figure
%         {
%             % \label{fig:1}
%             \includegraphics[width=0.45\textwidth, height=0.4\textwidth]{chapter1/figures/classifier_accuracy.png}
%         }%
%         \subfigure[]%Caption of Second Figure
%         {%
%           % \label{fig:2}
%           \includegraphics[width=0.45\textwidth, height=0.4\textwidth]{chapter1/figures/classifier_loss.png}
%         }\\ %  ------- End of the first row ----------------------%
        
%     \end{center}
%     \caption{%
%         \textbf{(a)} Classification accuracy for our modified DensNet-121, \textbf{(b)} Classification loss for our modified DensNet-121.}
%   \label{fig:accuracy_loss}
% \end{figure}

\begin{table}[!htbp] \centering
\caption{Comparison of Different Experiment Setup.}
\label{tab:comparison_exp}
\begin{tabular}{@{\extracolsep{5pt}} llrrrr} 
\\[-1.8ex]\hline 
\hline \\[-1.8ex] 
\multicolumn{1}{c}{Used Model} & \multicolumn{1}{c}{Top-1 Accuracy (\%)} & \multicolumn{1}{c}{Top-5 (\%)} & \multicolumn{1}{c}{Precision } & \multicolumn{1}{c}{Recall} & \multicolumn{1}{c}{F1-Score} \\
\hline \\[-1.8ex] 
\midrule
 	ResNet-50 & 77 & 94 & 0.81 & 0.76 & 0.76\\
  	ResNet-101 & 75 & 92 & 0.77 & 0.73 & 0.73\\
  	VGG-16 & 74 & 92 & 0.73 & 0.70 & 0.69\\
  	DenseNet-121 & 85 & 96 & 0.87 & 0.83 & 0.83\\
  	
\hline \\[-1.8ex] 
\end{tabular}
\end{table}

\begin{table}[!htbp] \centering
\caption{Comparison with other Methods}
\label{tab:comparison_methos}
\begin{tabular}{@{\extracolsep{5pt}} llrrrr} 
\\[-1.8ex]\hline 
\hline \\[-1.8ex] 
\multicolumn{1}{c}{Method} & \multicolumn{1}{c}{Used Model} & \multicolumn{1}{c}{Top-1 Accuracy (\%)} & \multicolumn{1}{c}{Top-5 (\%)} \\
\hline \\[-1.8ex] 
\midrule
  	\cite{15_chasmai2022view} & DCPose(Yoga-82) dataset & 77 & - \\
 	\cite{1_verma2020yoga} & DenseNet-201 & 74.91 & 91.30 \\
  	\cite{1_verma2020yoga} & ResNet-50 & 63.44 & 82.55 \\
  	
\hline \\[-1.8ex] 
\end{tabular}
\end{table}

\section{Conclusion and Future Work}
\label{sec_conclusion_and_future_work}
The main challenges to classifying yoga poses are that they have very repetitive poses and so they are very difficult to annotate as a body key point. That’s why the researchers have moved away from key point-based classification to an image classification problem. Due to advancements in deep learning and deep neural networks, these types of multiclass classification problems have become easier. In this work, we have explored the power of transfer learning, used pre-trained deep models, and modified them to serve our purposes. And it has shown promising results in terms of top-1 and top-5 accuracy.

We would like to explore more like ensembling deep learning architectures and different image processing techniques to move forward with the current research. We also want to investigate the explainability of the model using GradCam and similar types of algorithms. We believe this will help us to analyze errors in a more convenient way alongside handling any kind of biases.

\let\cleardoublepage\clearpage

% \bibliographystyle{plain}
%\bibliography{bibtex_example}
%\bibliography{references}

%\setcounter{chapter}{10}
%\include{chapter2/ch2}

% \bibliographystyle{plain}

% \bibliography{references}

%\printindex
%\cleardoublepage
\end{document}